\documentclass{article}

\usepackage{spconf,amsmath,graphicx}
\usepackage{xcolor}

\usepackage[T1]{fontenc}    
\usepackage{hyperref}       
\usepackage{url}            
\usepackage{booktabs}       
\usepackage{amsfonts}       
\usepackage{nicefrac}       
\usepackage{microtype}      
\usepackage{xcolor}         
\usepackage{booktabs}
\usepackage{multirow}
\usepackage{graphicx}

\usepackage{colortbl}
\usepackage[ruled]{algorithm2e}
\usepackage{amsmath}

\title{GTFMN: GUIDED TEXTURE AND FEATURE MODULATION NETWORK FOR LOW-LIGHT IMAGE ENHANCEMENT AND SUPER-RESOLUTION}

\name{\parbox{\textwidth}{\centering
      Yongsong Huang$^{1,2}$, 
      Tzu-Hsuan Peng$^3$, Tomo Miyazaki$^1$, Xiaofeng Liu$^2$, \\
      Chun-Ting Chou$^4$, Ai-Chun Pang$^{3,5}$, Fellow, IEEE, Shinichiro Omachi$^1$, Senior Member, IEEE
      }}

\address{$^1$ Graduate School of Engineering, Tohoku University, Sendai, Japan \\
         $^2$ Dept. of Radiology \& Biomedical Imaging, Yale University, New Haven, CT, USA \\
         $^3$ Dept. of Computer Science and Information Engineering, National Taiwan University, Taipei, Taiwan \\
         $^4$ Graduate Institute of Communication Engineering, National Taiwan University, Taipei, Taiwan \\
         $^5$ Research Center for Information Technology Innovation, Academia Sinica, Taipei, Taiwan}

\begin{document}
%
\maketitle
\begin{abstract}
Low-light image super-resolution (LLSR) is a challenging task due to the coupled degradation of low resolution and poor illumination. To address this, we propose the Guided Texture and Feature Modulation Network (GTFMN), a novel framework that decouples the LLSR task into two sub-problems: illumination estimation and texture restoration. First, our network employs a dedicated Illumination Stream whose purpose is to predict a spatially varying illumination map that accurately captures lighting distribution. Further, this map is utilized as an explicit guide within our novel Illumination Guided Modulation Block (IGM Block) to dynamically modulate features in the Texture Stream. This mechanism achieves spatially adaptive restoration, enabling the network to intensify enhancement in poorly lit regions while preserving details in well-exposed areas. Extensive experiments demonstrate that GTFMN achieves the best performance among competing methods on the OmniNormal5 and OmniNormal15 datasets, outperforming them in both quantitative metrics and visual quality.
\end{abstract}

\section{Introduction}
\label{sec:intro}

Image super-resolution (SR) and low-light enhancement are fundamental computer vision tasks with widespread applications in areas such as autonomous driving and video surveillance\cite{huang2021infrared,11059944,huang2022infrared}. In practical scenarios, particularly with hardware-constrained devices like vehicle dashcams, captured images often suffer from both low resolution and severe illumination degradation due to rapidly changing outdoor lighting\cite{zhuang2025local,zhao2024non}. For critical situations such as accident analysis, high-quality image restoration is paramount, necessitating the simultaneous execution of super-resolution and low-light enhancement. While a two-stage approach that handles these degradations sequentially is feasible, it inevitably introduces additional computational overhead. Therefore, developing a unified framework that addresses both issues concurrently is highly desirable.

\begin{figure}[t]
\centering
\includegraphics[width=\columnwidth]{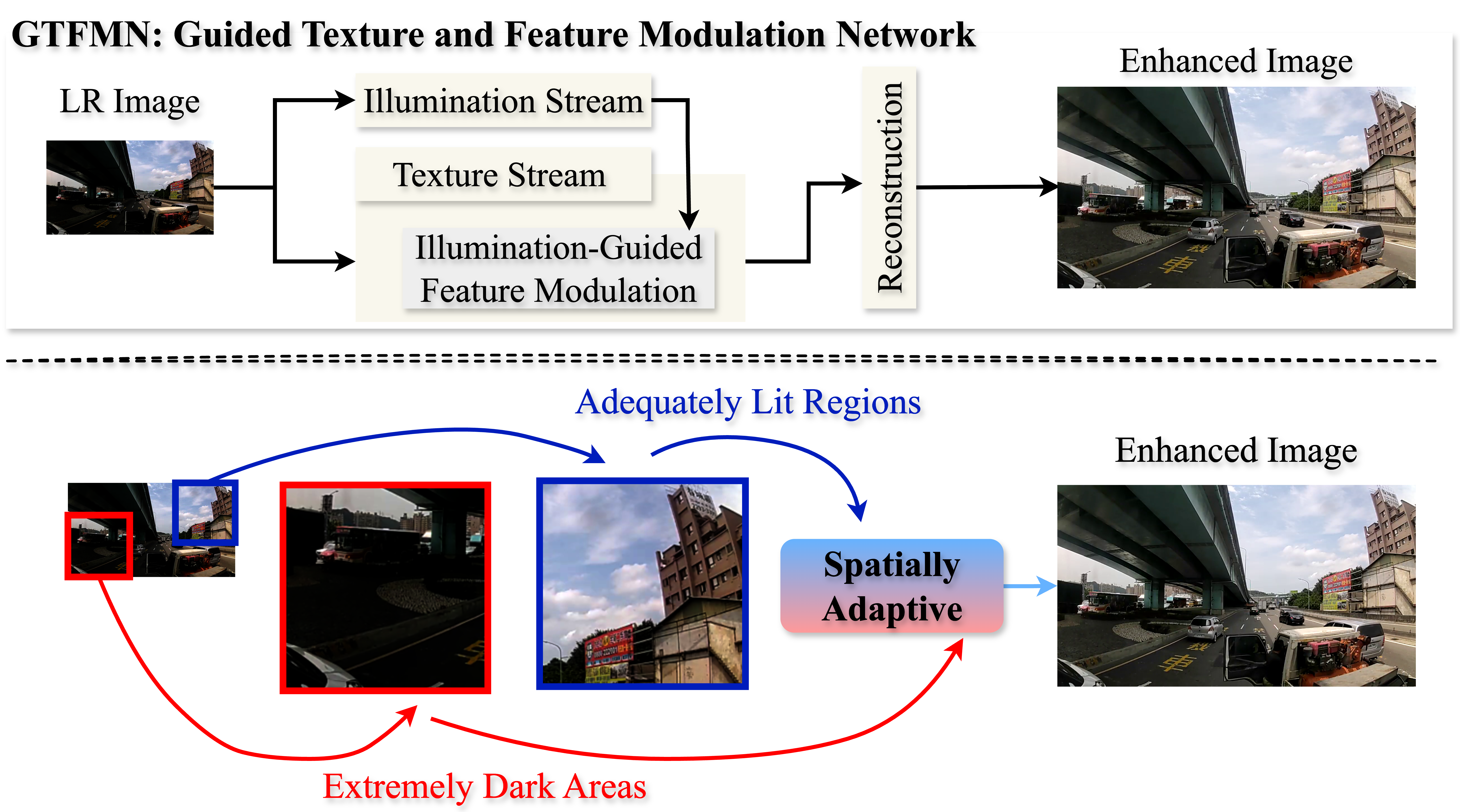}
\caption{The overall framework and core concept of GTFMN.}
\label{fig1}
\end{figure}

Conventional single-stream SR networks\cite{dong2015image,sun2023spatially} often falter in this challenging domain\cite{zhou2022domain,zhang2022transfer}. When applied directly to low-light images, they tend to amplify inherent noise and produce undesirable color shifts\cite{sun2016return}, as their feature extraction mechanisms are not designed to distinguish between signal and noise under such conditions. As illustrated in Fig.\ref{fig1}, the core challenge is to selectively enhance extremely dark areas without over-exposing regions that are already adequately lit, demanding a spatially adaptive solution\cite{cai2023retinexformer,bai2024retinexmamba,10.2312:pg.20231267}.

\begin{figure*}[htbp]
\centering
\includegraphics[width=\textwidth]{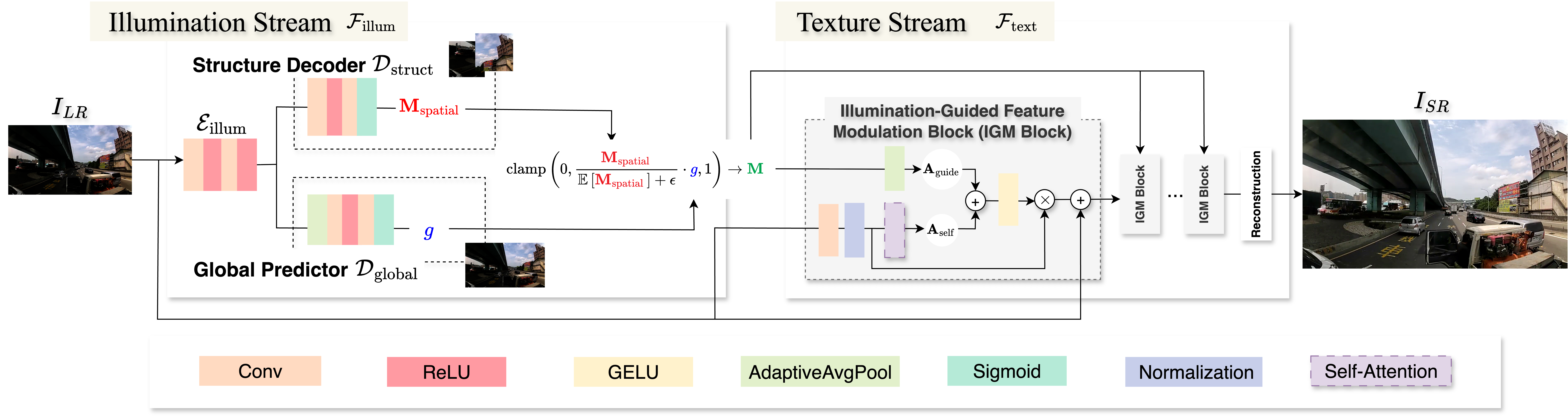}
\caption{The detailed architecture of our GTFMN. The network consists of two main branches. The \textbf{Illumination Stream} (left) takes the low-resolution input ($I_{LR}$) and employs a decoupled design with a Structure Decoder and a Global Predictor to estimate a stable illumination map ($\mathbf{M}$). The \textbf{Texture Stream} (right) processes the image features through a series of Illumination-Guided Feature Modulation Blocks (IGM Blocks). Within each IGM Block, the $\mathbf{M}$ generates a guided attention map ($\mathbf{A}_{\text {guide }}$), which is fused with the feature-derived self-attention map ($\mathbf{A}_{\text {self}}$) to dynamically modulate the texture features. Finally, a reconstruction module produces the super-resolution image ($I_{SR}$).
}
\label{fig2}
\end{figure*}

To address this, we propose a novel approach that explicitly decouples the problem. Instead of relying on a single network to implicitly learn both tasks, we introduce the Guided Texture and Feature Modulation Network (GTFMN, see Fig.\ref{fig1}). Specifically, our framework is designed to first analyze the scene's illumination and subsequently leverage this analysis to guide the texture restoration process. This is realized through a novel dual-stream architecture: \textbf{1)} An Illumination Stream is designed to estimate a pixel-wise illumination map from the low-resolution input. To enhance stability and physical plausibility, this stream further decouples the estimation of spatial light distribution and global brightness.  \textbf{2)} A Texture Stream processes the image content, where a series of novel Illumination-Guided Feature Modulation Blocks (IGM Blocks) leverage the estimated illumination map to dynamically modulate texture features. This allows the network to apply enhancement adaptively across different regions.

Our contributions are as follows: First, we propose a novel dual-stream architecture that decouples the LLSR task into illumination estimation and texture restoration. Second, we introduce the IGM Block, a new module that leverages the estimated illumination as an explicit guide to achieve spatially adaptive feature modulation. Third, we validate our method through extensive experiments, which demonstrate that GTFMN achieves a favorable balance between performance and parameter count.

\begin{figure*}[ht]
\centering
\includegraphics[width=\textwidth]{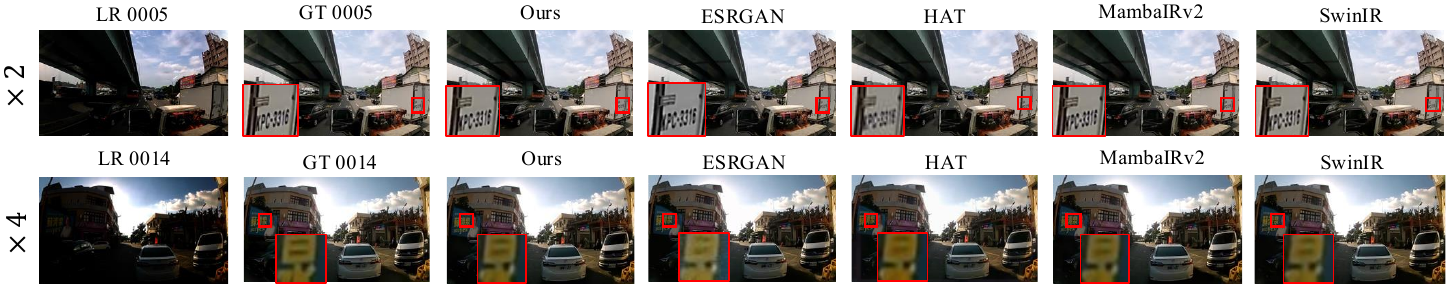}
\caption{Qualitative comparison with state-of-the-art methods on the OmniNormal15 dataset for $\times2$ and $\times4$ SR.}
\label{fig3}
\end{figure*}

\begin{table*}[ht]
\centering
\renewcommand\arraystretch{1.1}
\caption{The average results of (PSNR/dB$\uparrow$ MSE$\downarrow$ SSIM$\uparrow$ LPIPS$\downarrow$) with scale factor of 2 \& 4 on datasets OmniNormal5 \& OmniNormal15. Best and second-best performances are marked in \textbf{bold} and \underline{underlined}, respectively.}
\resizebox{\textwidth}{!}{%
\begin{tabular}{@{}c|c|c|cccc|cccc@{}}
\toprule
\multirow{2}{*}{Scale} & \multirow{2}{*}{Methods} & \multirow{2}{*}{\# Params. (K)} & \multicolumn{4}{c|}{OmniNormal5} & \multicolumn{4}{c}{OmniNormal15} \\ \cmidrule(l){4-11} 
 &  &  & PSNR/dB$\uparrow $ & MSE$\downarrow $ & SSIM$\uparrow $ & \multicolumn{1}{c|}{LPIPS$\downarrow $} & PSNR/dB$\uparrow $ & MSE$\downarrow $ & SSIM$\uparrow $ & LPIPS$\downarrow $ \\ \midrule
\multirow{10}{*}{$\times 2$} 
 & SRCNN\textcolor[RGB]{217,205,144}{\textit{[T-PAMI 2015]}}~\cite{dong2015image} & 57 & 34.9111 & 23.7764 & 0.9655 & 0.1327 & 34.9288 & 26.0693 & 0.9668 & 0.1243 \\
 & FSRCNN\textcolor[RGB]{217,205,144}{\textit{[ECCV 2016]}}~\cite{dong2016accelerating} & 475 & 36.9519 & 15.0543 & 0.9780 & 0.0986 & 37.1436 & 15.7841 & 0.9793 & 0.0898 \\
 & RCAN\textcolor[RGB]{217,205,144}{\textit{[ECCV 2018]}}~\cite{zhang2018image} & 12,467 & 25.1271 & 218.7846 & 0.9657 & 0.1264 & 24.3144 & 261.1591 & 0.9613 & 0.1307 \\
 & ESRGAN\textcolor[RGB]{217,205,144}{\textit{[ECCVW 2018]}}~\cite{wang2018esrgan} & 16,661 & \underline{38.1424} & \underline{11.1514} & \underline{0.9830} & \underline{0.0828} & \underline{38.1943} & \underline{11.8353} & \underline{0.9834} & \textbf{0.0748} \\
 & SwinIR\textcolor[RGB]{217,205,144}{\textit{[ICCV 2021]}}~\cite{liang2021swinir} & 11,752 & 37.2010 & 13.7718 & 0.9799 & 0.0948 & 37.5008 & 14.2757 & 0.9811 & 0.0878 \\
 & ShuffleMixer (base)\textcolor[RGB]{217,205,144}{\textit{[NIPS'22]}}~\cite{sun2022shufflemixer} & 121 & 37.0043 & 14.8764 & 0.9782 & 0.0973 & 37.2787 & 15.2924 & 0.9798 & 0.0890 \\
 & ShuffleMixer(tiny)\textcolor[RGB]{217,205,144}{\textit{[NIPS'22]}}~\cite{sun2022shufflemixer} & 108 & 37.4738 & 13.3795 & 0.9801 & 0.0928 & 37.6842 & 13.8223 & 0.9815 & 0.0842 \\
 & HAT\textcolor[RGB]{217,205,144}{\textit{[CVPR 2023]}}~\cite{Chen_2023_CVPR} & 20,624 & 37.1974 & 14.0158 & 0.9795 & 0.0942 & 37.3550 & 14.8403 & 0.9807 & 0.0880 \\
 & MambaIRv2 \textcolor[RGB]{217,205,144}{\textit{[CVPR 2025 SOTA]}}~\cite{guo2025mambairv2} & 22,903 & 37.9634 & 11.6950 & 0.9820 & 0.0842 & 38.1523 & 12.2329 & 0.9827 & 0.0793 \\
 & \textbf{Ours} & 8,784 & \textbf{38.3420} & \textbf{10.6568} & \textbf{0.9833} & \textbf{0.0802} & \textbf{38.4256} & \textbf{11.3635} & \textbf{0.9837} & \underline{0.0764} \\
 \midrule
\multirow{10}{*}{$\times 4$} 
 & SRCNN\textcolor[RGB]{217,205,144}{\textit{[T-PAMI 2015]}}~\cite{dong2015image} & 57 & 27.6026 & 119.0490 & 0.8523 & 0.2924 & 27.3339 & 141.8247 & 0.8478 & 0.2743 \\
 & FSRCNN\textcolor[RGB]{217,205,144}{\textit{[ECCV 2016]}}~\cite{dong2016accelerating} & 475 & 28.9793 & 87.0475 & 0.8912 & 0.2398 & 28.9108 & 100.9445 & 0.8912 & 0.2289 \\
 & RCAN\textcolor[RGB]{217,205,144}{\textit{[ECCV 2018]}}~\cite{zhang2018image} & 12,467 & 19.8145 & 692.1308 & 0.8109 & 0.3689 & 19.2857 & 790.8084 & 0.8082 & 0.3400 \\
 & ESRGAN\textcolor[RGB]{217,205,144}{\textit{[ECCVW 2018]}}~\cite{wang2018esrgan} & 16,661 & \underline{30.0527} & \underline{66.9510} & \underline{0.9149} & \underline{0.2143} & \underline{30.0691} & \underline{78.6757} & \underline{0.9129} & \underline{0.2032} \\
 & SwinIR\textcolor[RGB]{217,205,144}{\textit{[ICCV 2021]}}~\cite{liang2021swinir} & 11,752 & 28.7296 & 92.0697 & 0.8850 & 0.2459 & 28.6221 & 106.5887 & 0.8855 & 0.2338 \\
 & ShuffleMixer (base)\textcolor[RGB]{217,205,144}{\textit{[NIPS'22]}}~\cite{sun2022shufflemixer} & 121 & 29.7385 & 72.1371 & 0.9079 & 0.2171 & 29.6922 & 85.4311 & 0.9065 & 0.2077 \\
 & ShuffleMixer(tiny)\textcolor[RGB]{217,205,144}{\textit{[NIPS'22]}}~\cite{sun2022shufflemixer} & 108 & 29.2293 & 82.0615 & 0.8973 & 0.2319 & 29.2774 & 94.1718 & 0.8984 & 0.2200 \\
 & HAT\textcolor[RGB]{217,205,144}{\textit{[CVPR 2023]}}~\cite{Chen_2023_CVPR} & 20,624 & 28.5813 & 94.1024 & 0.8879 & 0.2506 & 28.2677 & 110.9638 & 0.8881 & 0.2435 \\
 & MambaIRv2 \textcolor[RGB]{217,205,144}{\textit{[CVPR 2025 SOTA]}}~\cite{guo2025mambairv2} & 22,903 & 29.8158 & 70.8697 & 0.9083 & 0.2196 & 29.7476 & 85.2837 & 0.9053 & 0.2097 \\
 & \textbf{Ours} & 8,784 & \textbf{31.1362} & \textbf{51.8216} & \textbf{0.9303} & \textbf{0.1859} & \textbf{30.5956} & \textbf{70.7727} & \textbf{0.9187} & \textbf{0.1890} \\
\bottomrule
\end{tabular}%
}
\label{tab.1}
\end{table*}

\section{Methodology}
\label{sec:methodology}

Our proposed GTFMN is designed to transform a low-light, low-resolution input image $I_{LR} \in \mathbb{R}^{H \times W \times 3}$ into a high-quality, normal-light output $I_{SR} \in \mathbb{R}^{sH \times sW \times 3}$, where $H$ and $W$ are the height and width of the input, and $s$ is the upscaling factor. The network, denoted as $\mathcal{G}$, also produces an intermediate illumination map $\mathbf{M} \in [0, 1]^{H \times W \times 1}$. The overall process is formulated as:
\begin{equation}
\left(I_{S R}, \mathbf{M}\right)=\mathcal{G}\left(I_{L R}\right)
\end{equation}
The architecture, depicted in Fig. \ref{fig2}, consists of two parallel streams followed by a feature modulation and reconstruction pipeline.

\subsection{Illumination Stream}

The Illumination Stream, $\mathcal{F}_{\text{illum}}$, is responsible for estimating the scene's lighting conditions. To achieve a robust estimation, we decouple this process into predicting spatial structure and global intensity. First, an encoder $\mathcal{E}_{\text{illum}}$ extracts features $\mathbf{F}_{\text{enc}}$ from the input $I_{LR}$. Subsequently, two parallel branches operate on these features: A Structure Decoder, $\mathcal{D}_{\text{struct}}$, predicts the spatial illumination distribution, yielding a preliminary map $\mathbf{M}_{\text{spatial}} \in [0, 1]^{H \times W \times 1}$. Further, a Global Predictor, $\mathcal{D}_{\text{global}}$, employs adaptive average pooling to estimate the scene's global mean brightness, producing a scalar value $g \in [0, 1]$.

The final illumination map $\mathbf{M}$ is synthesized by normalizing the spatial map and scaling it by the global intensity, ensuring its mean value aligns with the predicted global brightness. This is formally expressed as:
\begin{equation}
\mathbf{M} = \operatorname{clamp}\left(0, \frac{\mathbf{M}_{\text {spatial }}}{\mathbb{E}\left[\mathbf{M}_{\text {spatial }}\right]+\epsilon} \cdot g, 1\right)
\end{equation}
where $\mathbb{E}[\cdot]$ denotes the spatial mean operator and $\epsilon$ is a small constant to prevent division by zero. This decoupled design, e.g., GTFMN, enhances the stability and physical plausibility of the estimated illumination.

\subsection{Texture Stream and Guided Modulation}

The Texture Stream is the core of the restoration process, guided by the output of the Illumination Stream. Initially, a convolution layer maps the input $\mathbf{I}_{LR}$ into a high-dimensional feature space, producing the initial feature map $\mathbf{F}_0 \in \mathbb{R}^{H \times W \times C}$.

This feature map is then processed by a series of IGM Blocks. Each block, denoted $\mathcal{B}_i$, refines the feature map $\mathbf{F}_{i-1}$ using the illumination map $\mathbf{M}$ as guidance:
\begin{equation}
\mathbf{F}_i=\mathcal{B}_i\left(\mathbf{F}_{i-1}, \mathbf{M}\right), \quad \text { for } i=1, \ldots, N
\end{equation}
The central mechanism within each IGM Block is the fusion of self-derived and illumination-guided attention. For an input feature $\mathbf{F}_{\text{in}}$, we first compute a self-attention map $\mathbf{A}_{\text{self}}$ using a multi-scale attention layer. Next, the illumination map $\mathbf{M}$ is passed through a small adapter network to generate a guided attention map $\mathbf{A}_{\text{guide}}$. These two maps are additively fused to form the final modulation signal:
\begin{equation}
\mathbf{A}_{\text {final }}=\mathbf{A}_{\text {self }}+\mathbf{A}_{\text {guide }}
\end{equation}
This final attention map dynamically modulates the normalized input features through element-wise multiplication. The output of the IGM Block is produced after passing through a feed-forward network with residual connections.

Finally, the reconstruction layer $\mathcal{F}_{\text{recon}}$, composed of a convolution layer and a PixelShuffle operation, then upscales the deep features to generate $I_{SR}$.

\section{Experimental Results}
\label{sec:Experimental Results}

\textbf{Datasets.} Our experiments are conducted on a custom-built dataset. The training set, named OmniTrain, comprises 1600 HR images paired with their synthetically degraded counterparts. For evaluation, we utilize two distinct test sets: OmniNormal5 and OmniNormal15, which contain 5 and 15 representative scenes, respectively. These test datasets are designed to facilitate a fair and comprehensive comparison of algorithm performance. To generate the paired data, we first analyzed a large set of real-world low-light images to establish a representative average gamma value. We then applied this gamma correction to the high-quality images to simulate low-light conditions, followed by bicubic downsampling to create the final LR images. This process ensures our dataset realistically models the compound degradations encountered in practical applications.

\textbf{Metrics and Implementation Details.} We report performance using four standard metrics: Peak Signal-to-Noise Ratio (PSNR), Structural Similarity (SSIM), Mean Squared Error (MSE), and the Learned Perceptual Image Patch Similarity (LPIPS)\cite{snell2017learning} to measure perceptual quality. Our model is trained using the Adam\cite{kingma2014adam} optimizer with an L1 loss function. The number of IGM Blocks is set to 64. All evaluations are conducted in the $\textbf{Y}$ channel of the YCbCr color space. The learning rate is initialized at $2 \times 10^{-4}$. Utilizing the PyTorch framework, the model is trained with an RTX A6000 GPU.


\textbf{Quantitative Results.} As shown in Table.\ref{tab.1}, we compare GTFMN with several methods. Our model consistently outperforms all competing methods across both datasets and scaling factors ($\times2$, $\times4$). Notably, for the challenging $\times4$ super-resolution task on OmniNormal5, GTFMN achieves a PSNR of 31.1362 dB, surpassing the second-best method, ESRGAN, by a significant margin of 1.08 dB. Furthermore, GTFMN demonstrates strong parameter efficiency, achieving these results with only 8.78M parameters, which is considerably less than large-scale models like RCAN, SwinIR, and HAT.


\textbf{Qualitative Result.} Fig. 3 provides a visual comparison of our results against other state-of-the-art methods. Visual comparisons reveal that competing models like ESRGAN and HAT tend to amplify noise and generate artifacts, particularly in the darkest regions of the image (highlighted by red boxes). In contrast, GTFMN more effectively restores textures and suppresses noise, producing results with higher fidelity. For instance, in test sample "LR 0005", our method renders the text "KPC-3316" with superior clarity and sharpness compared to all other methods. Similarly, in the "LR 0014" sample, GTFMN restores the details on the yellow sign more faithfully.

\begin{figure}
\centering
\includegraphics[width=0.7\columnwidth]{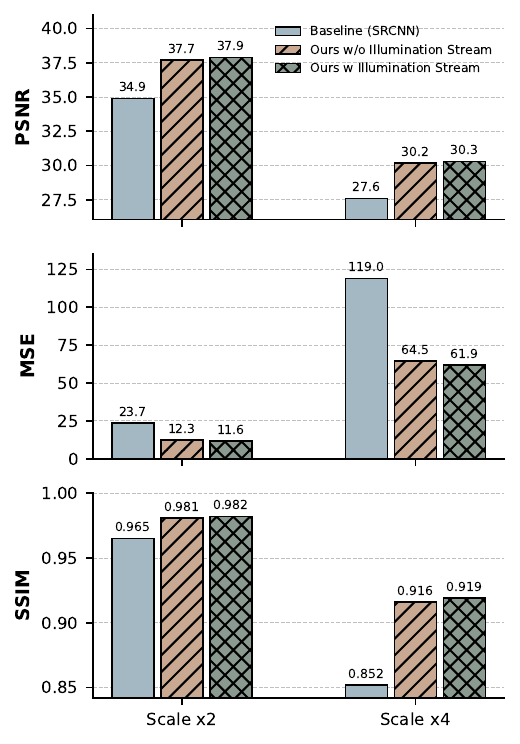}
\caption{Ablation study on the effectiveness of the Illumination Stream.}
\label{fig4}
\end{figure}

\subsection{Ablation Study}
\label{sec3.2}

\begin{table}[t]
\centering
\caption{Ablation study on the number of blocks. The best result for each dataset is highlighted in \textbf{bold}.}
\label{tab.2}

\setlength{\tabcolsep}{6pt} 

\begin{tabular}{c cc cc}
    \toprule
    \multirow{2}{*}{\# Blocks} & \multicolumn{2}{c}{OmniNormal5} & \multicolumn{2}{c}{OmniNormal15} \\
    \cmidrule(lr){2-3} \cmidrule(lr){4-5}
    & PSNR (dB) & SSIM & PSNR (dB) & SSIM \\
    \midrule
    16 & 38.0051 & 0.9820 & 38.0172 & 0.9821 \\
    32 & 37.9985 & 0.9820 & 38.2557 & 0.9831 \\
    64 & \textbf{38.1061} & \textbf{0.9824} & \textbf{38.3408}    & \textbf{0.9834}     \\
    \bottomrule
\end{tabular}
\end{table}

\textbf{Effectiveness of the IGM Blocks.} We experimented with varying the number of IGM Blocks in the Texture Stream. As shown in Table \ref{tab.2}, performance generally improves with network depth. The configuration with 64 blocks achieves the best results on both datasets, demonstrating the capacity of a deeper network to learn more complex feature transformations. However, it is important to note that a deeper network also leads to an increase in parameter count and computational overhead. This presents a trade-off, allowing the network depth to be selected based on specific application requirements. 

\textbf{Effectiveness of the Illumination Stream.} We also compare our full model against a variant where the Illumination Stream and all guidance mechanisms are removed, reducing it to a single-stream architecture. As shown in Fig.\ref{fig4}, the full model ("Ours w Illumination Stream") largely outperforms the variant without guidance ("Ours w/o Illumination Stream") across all metrics. For instance, at scale $\times4$, the inclusion of the illumination stream boosts PSNR from 30.2 dB to 30.3 dB and improves SSIM from 0.916 to 0.919. This confirms that the explicit illumination guidance is crucial for achieving high-quality restoration.

\section{Conclusion}

In this paper, we introduced GTFMN, a novel dual-stream network for low-light image super-resolution. By decoupling the task into illumination estimation and guided texture restoration, our model effectively addresses the coupled degradations inherent in the LLSR problem. The proposed Illumination-Guided Modulation Block enables spatially adaptive feature enhancement, leading to better performance in both quantitative metrics and visual quality. Our work provides a new and effective framework for tackling complex, multi-faceted image restoration tasks.


\section{Acknowledgement}
This work was supported in part by JSPS KAKENHI Grant (JP23KJ0118, JP23K11176, and JP25K03130), the National Science and Technology Council Grant No.NSTC112-2221-E-001-028-MY3, the NVIDIA Academic Grant Program, and the Center of Data Intelligence: Technologies, Applications, and Systems, National Taiwan University (Grant No.115L900903), from the Featured Areas Research Center Program within the framework of the Higher Education Sprout Project by the Ministry of Education of Taiwan.

\bibliographystyle{IEEEbib}
\bibliography{strings,icassp26}

\end{document}